\pdfoutput=1
\documentclass[11pt]{article}

\usepackage[final]{acl}
\usepackage{natbib}
\usepackage{times}
\usepackage{makecell} 
\usepackage{amsmath}
\usepackage{amsfonts}
\usepackage{booktabs}
\usepackage{multirow}
\usepackage{subcaption}
\usepackage{latexsym}
\usepackage{booktabs}
\usepackage{fancyvrb}
\usepackage{enumitem}
\usepackage{multirow}
\usepackage{float}
\usepackage{algorithmic}
\usepackage{multirow}
\usepackage[T1]{fontenc}
\usepackage[utf8]{inputenc}

\usepackage{microtype}

\usepackage{inconsolata}

\usepackage{graphicx}
\usepackage{xcolor}  
\usepackage[utf8]{inputenc} % allow utf-8 input
\usepackage[T1]{fontenc}    % use 8-bit T1 fonts
\usepackage{hyperref}       % hyperlinks
\usepackage{url}            % simple URL typesetting
\usepackage{amsfonts}       % blackboard math symbols
\usepackage{nicefrac}       % compact symbols for 1/2, etc.
\usepackage{microtype}      % microtypography
\usepackage{graphicx}
\usepackage[most]{tcolorbox}
\usepackage{colortbl}
\usepackage{listings}

\lstdefinestyle{pythonstyle}{
    backgroundcolor=\color{gray!15},   % Set background color
    commentstyle=\color{blue},          % Set comment color
    keywordstyle=\color{magenta},         % Set keyword color
    numberstyle=\tiny\color{gray},       % Set line number style
    numbers=left,                        % Position of line numbers
    stepnumber=1,                        % Step size for line numbers
    numbersep=5pt,                       % Space between line numbers and code
    stringstyle=\color{red},             % Set string color
    basicstyle=\ttfamily\footnotesize,   % Basic font style
    breaklines=true,                     % Automatically break lines
    captionpos=b,                        % Position of the caption
    escapeinside={\%*}{*)}               % Escape inside comments
}

\usepackage[linesnumbered,ruled,vlined]{algorithm2e}

\author{
Yingxu Wang$^{1}$ \quad
Jiaxin Huang$^{1}$ \quad
Mengzhu Wang$^{2}$ \quad
Nan Yin$^{3}$ \\
$^{1}$Mohamed bin Zayed University of Artificial Intelligence \quad
$^{2}$Hebei University of Technology \\
$^{3}$City University of Hong Kong \\
\texttt{yingxv.wang@gmail.com, Jiaxin.Huang@mbzuai.ac.ae, dreamkily@gmail.com} \\
\texttt{yinnan8911@gmail.com}
}

\title{\method{}: An Experiential Framework for Coherent Multi-hop \\ Knowledge Graph Question Answering}

\def \method{TRACE}

\begin{document}
\maketitle
\begin{abstract}

% Multi-hop Knowledge Graph Question Answering (KGQA) requires coherent reasoning across relational paths, yet existing methods often treat each reasoning step independently and discard failed trajectories, resulting in fragmented and error-prone reasoning. To address these limitations, we propose \textbf{T}rajectory-aware \textbf{R}easoning with \textbf{A}daptive \textbf{C}ontext and \textbf{E}rror feedback (\method{}), an experiential framework that unifies LLM-driven contextual reasoning with failure-aware knowledge integration to enhance both the coherence and robustness of multi-hop KGQA. Specifically, \method{} dynamically translates evolving reasoning paths into natural language narratives to maintain semantic continuity, while systematically abstracting failed trajectories into reusable error patterns that capture recurring reasoning flaws. A dual-feedback re-ranking mechanism then integrates contextual narratives with distilled failure patterns to guide relation selection at each step. To validate the effectiveness of \method{}, we conduct comprehensive experiments on multiple KGQA benchmarks, demonstrating that \method{} consistently outperforms state-of-the-art baselines.

Multi-hop Knowledge Graph Question Answering (KGQA) requires coherent reasoning across relational paths, yet existing methods often treat each reasoning step independently and fail to effectively leverage experience from prior explorations, leading to fragmented reasoning and redundant exploration. To address these challenges, we propose \textbf{T}rajectory-aware \textbf{R}easoning with \textbf{A}daptive \textbf{C}ontext and \textbf{E}xploration priors (\method{}), an experiential framework that unifies LLM-driven contextual reasoning with exploration prior integration to enhance the coherence and robustness of multi-hop KGQA. Specifically, \method{} dynamically translates evolving reasoning paths into natural language narratives to maintain semantic continuity, while abstracting prior exploration trajectories into reusable experiential priors that capture recurring exploration patterns. A dual-feedback re-ranking mechanism further integrates contextual narratives with exploration priors to guide relation selection during reasoning. Extensive experiments on multiple KGQA benchmarks demonstrate that \method{} consistently outperforms state-of-the-art baselines.
\end{abstract}

\section{Introduction}

Knowledge Graph Question Answering (KGQA) aims to enable large language models (LLMs) to answer natural language questions by reasoning over structured knowledge graphs (KGs)~\cite{choi2023nutrea,saxena2020improving}. This task is crucial for bridging human language and symbolic knowledge, supporting a wide range of intelligent applications such as search engines~\cite{Zhao2020BrainInspiredSE,Kejriwal2017KnowledgeGF}, recommender systems~\cite{Wang2019KnowledgeGC,Guo2020ASO}, and personal assistants~\cite{Balog2019PersonalKG,Liu2024AQA,fang2025comprehensive}. Despite its importance, answering complex questions remains challenging since it often requires multi-hop reasoning over relational paths in KGs, demanding both accurate symbolic traversal and robust semantic understanding. 
% Recent advances in LLMs have opened a promising direction by leveraging their reasoning abilities to plan and execute relation paths~\cite{shen2025reasoning,ma2025debate,dammu2025dynamic}, yet effectively aligning their implicit reasoning with explicit KG structures remains an open problem.

% Knowledge Graphs (KGs) encode factual knowledge in a structured graph format and serve as the backbone of many intelligent applications such as search engines~\cite{Zhao2020BrainInspiredSE,Kejriwal2017KnowledgeGF}, recommender systems~\cite{Wang2019KnowledgeGC,Guo2020ASO}, and personal assistants~\cite{Balog2019PersonalKG,Liu2024AQA}. Knowledge Graph Question Answering (KGQA) connects natural language with these structured repositories, enabling users to retrieve information through intuitive queries~\cite{choi2023nutrea,saxena2020improving}. Answering complex questions often requires multi-hop reasoning, where a sequence of relations in the KG is traversed to link a topic entity to its answer. Recent progress in large language models (LLMs) has introduced a promising paradigm that leverages their strong reasoning abilities~\cite{shen2025reasoning,ma2025debate}. In this paradigm, LLMs navigate the symbolic space of KGs by planning and executing relation paths to identify answers, thereby enhancing both the transparency and flexibility of the reasoning process~\cite{dammu2025dynamic}.

In recent years, growing attention has been devoted to leveraging the reasoning capabilities of LLMs for KGQA~\cite{yu2022decaf,liu2025dual}. Existing studies vary in their assumptions about the roles LLMs should play within the reasoning pipeline, which can be broadly categorized into three paradigms. (1) LLMs as direct reasoners: the model is prompted with a natural language question and a serialized KG subgraph to infer answers in a zero-/few-shot manner\cite{sun2023think,bi2024forest}. (2) LLMs as stepwise explorers: inspired by ReAct~\cite{Yao2022ReActSR} and CoT~\cite{wei2022chain}, the model iteratively selects relations, traverses entities, and conditions future steps on prior results~\cite{wang2025dynamically,shen2025reasoning}. (3) LLMs as re-rankers: traditional graph search produces candidate paths that the LLM scores for semantic and logical coherence~\cite{liu2025dual,yao2025learning}. These paradigms demonstrate LLMs’ versatility in bridging symbolic and semantic reasoning~\cite{saxena2020improving,zhang2022subgraph}.
However, current methods still fail to integrate contextual information from previous reasoning steps or leverage experience from prior explorations, often resulting in incoherent reasoning paths and redundant explorations~\cite{shen2025reasoning}.

This paper investigates the design of an experiential KGQA framework that enhances coherent multi-hop reasoning and learns from prior explorations. However, realizing such a framework presents three core challenges. (1) Contextual guidance: existing KGQA models often treat each reasoning step independently, neglecting how previously traversed relations influence subsequent decisions~\cite{wang2025dynamically,shen2025reasoning}. This disrupts reasoning coherence and causes the reasoning path to deviate from the question’s intent. The challenge lies in dynamically translating structured relation paths into coherent natural language contexts that guide each reasoning step. (2) Exploration generalization: most approaches lack explicit mechanisms to incorporate prior exploration trajectories into subsequent reasoning steps, which limits their ability to avoid redundant explorations~\cite{ma2025debate,yao2025learning}. The key is to autonomously identify and abstract common patterns from historical explorations into generalizable priors. (3) Unified decision-making: experiential reasoning requires integrating the evolving reasoning context with accumulated exploration priors. Yet, existing methods lack mechanisms to synthesize these complementary sources of information~\cite{liu2025dual,ma2025large}. The challenge is to build a unified framework that fuses contextual cues and experiential priors at each step, ensuring coherent and robust multi-hop reasoning.

To address these challenges, we propose \textbf{T}rajectory-aware \textbf{R}easoning with \textbf{A}daptive \textbf{C}ontext and \textbf{E}xploration priors (\method{}), an experiential framework that unifies LLM-driven contextual reasoning with exploration-aware knowledge integration. \method{} dynamically translates evolving reasoning trajectories into natural language narratives, ensuring that each decision remains contextually grounded in both the input question and the accumulated reasoning history. Moreover, it abstracts prior exploration trajectories into reusable experiential priors, enabling the model to reduce redundant exploration and guide subsequent reasoning. In addition, a dual-feedback re-ranking mechanism integrates contextual narratives with exploration priors to refine relation selection, resulting in more coherent and robust multi-hop reasoning. Our contributions are summarized as follows:

% To address these challenges, we propose Multi-Prompt Contextual Reasoning (\method{}), an experiential framework that unifies LLM-driven contextual modeling with failure-aware knowledge integration. \method{} consists of three complementary components. First, we introduce a dynamic context generation module that translates evolving reasoning trajectories into natural language narratives, ensuring that each decision remains contextually grounded in the question and preceding steps. Second, we develop a failure generalization module that analyzes unsuccessful reasoning paths and distills them into reusable error patterns, allowing the model to avoid repeating similar mistakes. Third, we propose a dual-feedback re-ranking mechanism that first generates a set of top-$k$ candidate relations using an LLM-based retriever conditioned on the contextual narrative, and then refines their ranking by integrating the relation sequence of the current path with distilled failure patterns, ensuring robust and coherent relation selection. Our contributions are summarized as follows:

\begin{itemize}[itemsep=2pt,topsep=0pt,parsep=0pt,leftmargin=*]
 \item We investigate experiential path reasoning in KGQA, where the key challenges are to preserve the coherence of multi-hop reasoning paths and to leverage experience from prior explorations, motivating the development of context-aware and exploration-guided reasoning strategies.

 \item We propose \method{}, a novel framework that generates context-rich narratives during reasoning and abstracts experience from prior explorations into reusable priors, enabling robust and coherent multi-hop reasoning over knowledge graphs.

 \item We conduct comprehensive experiments on multiple KGQA benchmarks, showing that the proposed \method{} consistently outperforms state-of-the-art baselines.  
\end{itemize}

\section{Related Work}

\paragraph{LLMs for KGQA.} 

The integration of LLMs into KGQA has advanced rapidly in recent years. One line of research employs a retrieve-then-read paradigm, wherein a relevant subgraph of the knowledge graph is first retrieved, serialized, and then provided to an LLM to generate the answer~\cite{yao2025learning,liu2025dual}. While effective in certain scenarios, these methods inherently separate the retrieval and reasoning processes, thereby constraining the LLM’s capacity to adaptively refine path exploration based on intermediate reasoning outcomes~\cite{ma2025debate,fang2024karpa}. In contrast, another line of work enables LLMs to interact with the knowledge graph in a sequential, stepwise manner, allowing the model to iteratively select relations to traverse and explicitly construct multi-hop reasoning paths~\cite{sun2023think,ma2025deliberation,ma2024think}. However, these methods still suffer from insufficient incorporation of reasoning path context, as relation selection at each step is typically performed without consideration of the evolving semantic narrative of the reasoning process~\cite{wang2025dynamically,shen2025reasoning,lee2024zero}. This often leads to logical inconsistencies and deviations from the original question intent. To mitigate this limitation, we introduce \method{}, a novel framework that incorporates natural language narratives at every step of the reasoning process, enabling the model to capture evolving semantics and produce more coherent and accurate multi-hop KGQA.

\begin{figure*}[t!]
  \centering
\includegraphics[width=\textwidth]{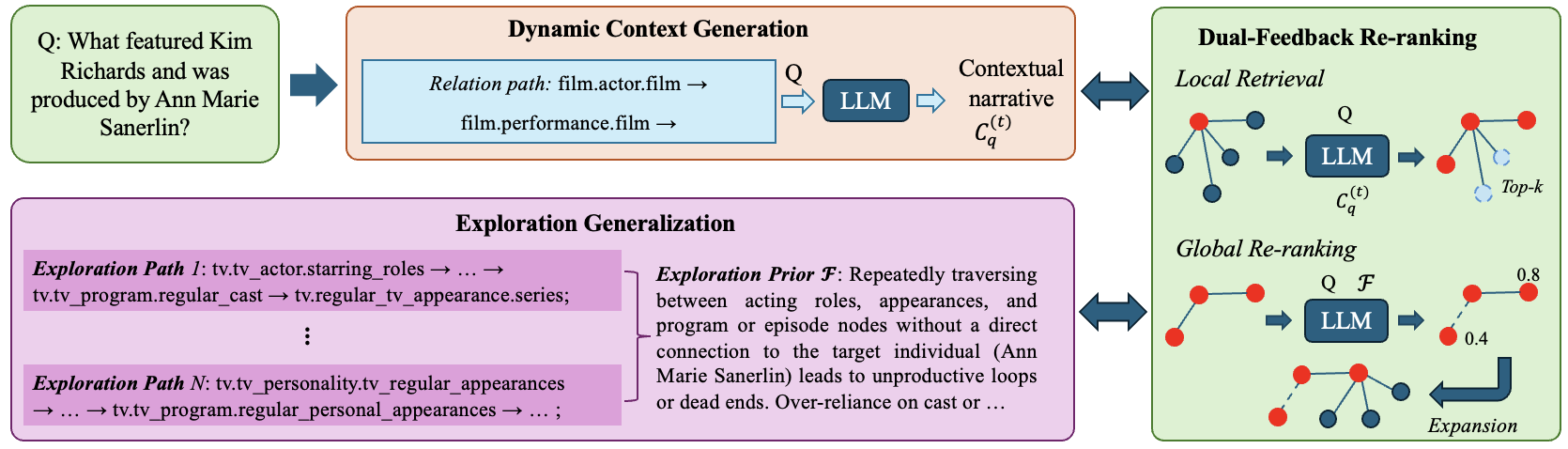} 
  % \vspace{-0.1cm}
\caption{Overview of the proposed \method{}.
Dynamic Context Generation translates evolving reasoning paths into natural language narratives, preserving semantic coherence across reasoning steps. Exploration Generalization abstracts prior exploration trajectories into reusable experiential priors that capture recurring exploration patterns. Finally, Dual-Feedback Re-ranking integrates contextual narratives with exploration priors to refine relation selection and enhance the robustness of multi-hop reasoning.}
  \label{framework}

\end{figure*}

\paragraph{Learning from Experience in Reasoning Models.} The ability to leverage experience from prior reasoning processes is crucial for developing robust reasoning systems. In pathfinding tasks, Reinforcement Learning (RL) has traditionally been employed, where sparse rewards are provided only upon successfully reaching a target~\citep{xiong2017deeppath,das2018go,wang2025evoagentx}. However, in complex multi-hop KGQA scenarios, successful reasoning paths are relatively rare, resulting in weak supervision signals~\cite{wang2025dynamically,zhang2025collaborative,liu2025sew}. As a consequence, acquiring effective exploration strategies under such limited feedback remains challenging. Recent studies have explored leveraging intermediate reasoning trajectories and historical exploration processes to improve reasoning efficiency and stability, but they lack principled mechanisms for abstracting recurring exploration patterns into reusable priors to accumulate strategic experience~\citep{wu2024self,pan2023automatically,huang2023large}. To address this limitation, we propose a framework that systematically distills generalizable exploration priors from historical reasoning trajectories, enabling the model to integrate accumulated experience and enhance robustness of multi-hop KGQA. 
\section{Methodology}

\subsection{Problem Formulation}

We formulate multi-hop Knowledge Graph Question Answering (KGQA) as a sequential decision-making problem over a knowledge graph $\mathcal{K} = {(e_s, r, e_o)} \subseteq \mathcal{E} \times \mathcal{R} \times \mathcal{E}$, where $\mathcal{E}$ and $\mathcal{R}$ denote the sets of entities and relations, respectively. Given a natural language question $q$ with an associated topic entity set $\mathcal{E}_q \subset \mathcal{E}$, the objective is to identify the correct set of answer entities $\mathcal{A}_q \subset \mathcal{E}$. This is achieved by discovering an optimal reasoning path $\mathcal{P}_q \subseteq \{(e_1, r_1, e_2), (e_2, r_2, e_3), \ldots\}$ such that traversing the path from a topic entity in $\mathcal{E}_q$ leads to the correct answer entities in $\mathcal{A}_q$.

\subsection{Overview of Framework}

In this paper, we propose \textbf{Trajectory-aware Reasoning with Adaptive Context and Exploration Priors} (\method{}), an experiential framework designed to enhance the coherence and robustness of multi-hop KGQA. \method{} consists of three core components: (1) \textbf{Dynamic Context Generation.} At each reasoning step, the evolving reasoning path is dynamically translated into a natural language narrative, ensuring that relation selection is informed by the accumulated reasoning context rather than treated as an isolated decision. (2) \textbf{Exploration Generalization.} When a reasoning trajectory terminates, \method{} summarizes the explored path to capture the semantic and structural decisions made during exploration. These summaries are periodically consolidated into generalizable exploration priors, which are then leveraged to inform subsequent reasoning steps. (3) \textbf{Dual-Feedback Re-ranking.} At each step, an LLM-based retriever generates a top-$k$ set of candidate relations conditioned on the contextual narrative, and their ranking is subsequently refined by incorporating the relation sequence of the current path together with the exploration priors.

\subsection{Dynamic Context Generation}

A central challenge in multi-hop KGQA is that relation selection at each step is often made in isolation, overlooking the evolving reasoning context. To overcome this limitation, the \textbf{Dynamic Context Generation} module translates preceding constructed reasoning paths into natural language narratives, which serve as contextual guidance for subsequent steps.

Formally, given a question $q$, we denote the preceding reasoning path up to step $t$ as
\begin{equation}
\mathcal{P}_q^{(t)} = \{(e_1, r_1, e_2), (e_2, r_2, e_3), \ldots, (e_t, r_t, e_{t+1})\},
\end{equation}
where $(e_i, r_i, e_{i+1}) \in \mathcal{K}$.  
The corresponding relation sequence is
\begin{equation}
R_q^{(t)} = (r_1, r_2, \ldots, r_t).
\end{equation}
At each step $t$, the relation sequence $R_q^{(t)}$ is transformed into a contextual narrative $C_q^{(t)}$ by an LLM-based generator $f_{\text{ctx}}(\cdot)$:
\begin{equation}
C_q^{(t)} = f_{\text{ctx}}(q, R_q^{(t)}),
\end{equation}
where $C_q^{(t)}$ is a natural language description of $R_q^{(t)}$ (e.g., ``find the director’s birthplace’’) that conveys the semantics of the traversed relations in the context of the question. 

By conditioning each decision on both the input question $q$ and the contextual narrative $C_q^{(t)}$, this module maintains semantic coherence across reasoning steps and reduces the risk of logical drift during multi-hop reasoning in KGQA.

\subsection{Exploration Generalization}\label{sec:eg}

A key limitation of existing KGQA approaches is that prior exploration trajectories are often not explicitly leveraged to inform subsequent reasoning, resulting in redundant exploration behaviors. To address this limitation, the \textbf{Exploration Generalization} module systematically summarizes prior exploration trajectories and abstracts them into reusable exploration priors. 

For a given question $q$, a reasoning trajectory $\mathcal{P}_q^{(t)}$ is considered a terminated exploration $\mathcal{T}$ if its associated relation sequence $R_q^{(t)}$ reaches the maximum step limit $L$ or cannot be further expanded:
\begin{equation}
\mathcal{T} = \mathcal{T}_{\text{depth}} \cup \mathcal{T}_{\text{expand}},
\end{equation}
where
\begin{equation}
\mathcal{T}_{\text{depth}} = \left\{ \mathcal{P}_q^{(t)} \;\middle|\; |R_q^{(t)}| \geq L \right\},
\end{equation}
and
\begin{equation}
\mathcal{T}_{\text{expand}} = \left\{ \mathcal{P}_q^{(t)} \mid \text{no expansion relations at step } t \right\},
\end{equation}
where $|R_q^{(t)}|$ denotes the number of traversed relations. For each terminated trajectory $\mathcal{P}_q^{(t)}$, an LLM-based summarization function $f_{\text{sum}}(\cdot)$ generates a trajectory summary $D_q$ that characterizes its relation sequence $R_q^{(t)}$:

\begin{equation}
D_q = f_{\text{sum}}(q, R_q^{(t)}).
\end{equation}

Although each trajectory summary $D_q$ captures useful information about a specific exploration process, it is inherently path-specific and lacks generality. To address this limitation, \method{} further aggregates and abstracts these trajectory summaries into a set of reusable exploration priors $\mathcal{F}$:
\begin{equation}
\mathcal{F} = f_{\text{gen}}\!\left(\{ D_q \;\middle|\; \mathcal{P}_q^{(t)} \in \mathcal{T} \}\right),
\end{equation}
where $f_{\text{gen}}(\cdot)$ is an abstraction function that identifies recurring exploration patterns. The resulting exploration priors $\mathcal{F}$ are stored as experiential knowledge and integrated into subsequent reasoning process in Sec.~\ref{sec:dual_feedback}.

\subsection{Dual-Feedback Re-ranking}
\label{sec:dual_feedback}
While Dynamic Context Generation provides local contextual narratives $C_q^{(t)}$ and Exploration Generalization yields global exploration priors $\mathcal{F}$, an effective mechanism is required to integrate these complementary signals during relation selection. To this end, \method{} introduces a \textbf{Dual-Feedback Re-ranking} module that combines both local and global guidance in a two-stage process.

At each step $t$, an LLM-based retriever $f_{\text{ret}}(\cdot)$ first generates a top-$k$ set of candidate relations from the neighborhood of the current entity $e_t$, conditioned on the contextual narrative $C_q^{(t)}$:
\begin{equation}
\mathcal{C}_t = f_{\text{ret}}(q, C_q^{(t)}, \mathcal{N}(e_t)),
\end{equation}
where $\mathcal{N}(e_t)$ denotes the set of outgoing relations associated with $e_t \in \mathcal{E}$.  
Subsequently, each candidate relation $r \in \mathcal{C}_t$ is re-ranked by an LLM-based scoring function $f_{\text{rank}}(\cdot)$, which is conditioned on the relation sequence $R_q^{(t)}$ and the exploration priors $\mathcal{F}$:
\begin{equation}
s(r) = f_{\text{rank}}(q, R_q^{(t)}, r, \mathcal{F}),
\end{equation}
where $s(r)$ denotes the relevance score assigned to relation $r$, reflecting the combined influence of local contextual information and exploration priors. 

Rather than restricting expansion to the single top-ranked relation, \method{} preserves all candidates whose scores exceed a confidence threshold $\zeta$, which is introduced as a hyperparameter. This design choice enables the framework to balance exploratory breadth with selection robustness. Each retained relation induces a new branch by extending the current path, resulting in a set of expanded reasoning trajectories:
\begin{equation}
R_q^{(t+1)} =
\left ( R_q^{(t)} \oplus r \;\middle|\; r \in \mathcal{C}_t,\ s(r) \geq \zeta \right ).
\end{equation}
where $\oplus$ denotes the operation of appending relation $r$ to the current relation sequence $R_q^{(t)}$.

By employing $C_q^{(t)}$ for candidate retrieval and $(R_q^{(t)}, \mathcal{F})$ for re-ranking, this module effectively integrates contextual semantics with experiential knowledge, thereby ensuring coherent and robust reasoning.

\subsection{Reasoning Process}

The overall reasoning pipeline of \method{} is illustrated in Appendix~\ref{sec:algorithm}. Reasoning trajectories are initialized from the topic entity set $\mathcal{E}_q$ and iteratively expanded through the joint use of Dynamic Context Generation and Dual-Feedback Re-ranking. At each step, the partial relation sequence $R_q^{(t)}$ is transformed into a contextual narrative $C_q^{(t)}$, which conditions both the retrieval and re-ranking of candidate relations. A trajectory $\mathcal{P}_q^{(t)}$ terminates when the maximum hop limit is reached or no feasible expansion is available. For each terminated trajectory, the Exploration Generalization module summarizes the explored path and abstracts it into exploration priors $\mathcal{F}$. Throughout the process, candidate relations with scores exceeding the confidence threshold $\zeta$ are retained to expand new reasoning trajectories. After finishing the above process, the answer set $\mathcal{A}$ is generated from the reasoning path with the highest relevance score.

\section{Experiments}

\subsection{Experimental Settings}

\begin{table}[t]
\centering
\label{tab:dataset_statistics}
\begin{tabular}{lccc}
\toprule
\textbf{Datasets} & \textbf{\#Train} & \textbf{\#Valid} & \textbf{\#Test} \\
\midrule
WebQSP     & 2,848   & 250    & 1,639  \\
\midrule
CWQ        & 27,639  & 3,519  & 3,531   \\
% \midrule
% MetaQA & 329,282 & 39,136 & 30,903 \\ 
\bottomrule
\end{tabular}
\caption{Statistics of KGQA benchmarks.}

\label{tab:datasets}
\end{table}

\paragraph{Datasets.}
To evaluate the effectiveness of \method{}, we conduct experiments on two widely used KGQA benchmarks: WebQSP~\cite{talmor2018web} and CWQ~\cite{yih2016value}. The statistics of these benchmarks are reported in Table~\ref{tab:datasets}. More details about datasets are provided in Appendix~\ref{sec:dataset}.

\paragraph{Baselines.} We compare \method{} with a comprehensive set of baselines. These baselines include: the semantic parsing methods, e.g., KV-Mem~\citep{miller2016key}, EmbedKGQA~\citep{saxena2020improving}, QGG~\citep{lan2020query}, NSM~\citep{he2021improving}, TransferNet~\citep{shi2021transfernet}, KGT5~\citep{saxena2022sequence}, and DECAF~\citep{yu2022decaf};
the retrieval-based methods, e.g., GraftNet~\citep{sun2018open}, PullNet~\citep{sun2019pullnet}, SR+NSM~\citep{zhang2022subgraph}, and SR+NSM+E2E~\citep{zhang2022subgraph};
the general LLMs, including 
Flan-T5-xl~\citep{chung2024scaling}, Alpaca-7B~\citep{taori2023stanford},
Llama3-8B~\citep{dubey2024llama}, Qwen2.5-7B~\citep{team2024qwen2}, ChatGPT~\citep{schulman2022chatgpt}, and ChatGPT+CoT~\citep{wei2022chain};
and recent LLMs with KG methods, including UniKGQA~\citep{jiang2022unikgqa}, KD-CoT~\citep{wang2023knowledge},
Nutrea~\citep{choi2023nutrea}, ToG~\citep{sun2023think}, RoG~\citep{luo2023reasoning}, KAPING~\citep{baek2023knowledge}, ReasoningLM~\citep{jiang2023reasoninglm},
FiDeLis~\citep{sui2024fidelis}, GNN-RAG~\citep{mavromatis2024gnn},
DoG~\citep{ma2025debate},  DualR~\citep{liu2025dual}
, DP~\citep{ma2025deliberation}, and RwT~\citep{shen2025reasoning}. The details are provided in Appendix~\ref{sec:baseline}.

\begin{table}[t]
\centering
\small
\resizebox{0.485\textwidth}{!}{
\begin{tabular}{clcccc}
\toprule
\multirow{2}{*}{\raisebox{-0.3\totalheight}{\textbf{{Type}}}} 
& \multirow{2}{*}{\raisebox{-0.3\totalheight}{\textbf{Methods}}} 
& \multicolumn{2}{c}{\textbf{WebQSP}} 
& \multicolumn{2}{c}{\textbf{CWQ}} \\
& & Hits@1 & F1 & Hits@1 & F1 \\
\midrule
\multirow{6}{*}{\rotatebox{90}{\makecell[c]{Semantic\\Parsing}}} 
& KV-Mem & 46.7 & 34.5 & 18.4 & 15.7 \\
& EmbedKGQA & 66.6 & - & 45.9 & - \\
& QGG & 73.0 & 73.8 & 36.9 & 37.4 \\
& NSM & 68.7 & 62.8 & 47.6 & 42.4 \\
& TransferNet & 71.4 & - & 48.6 & - \\
& KGT5 & 56.1 & - & 36.5 & - \\
& DECAF & 82.1 & 78.8 & 70.4 & - \\
\midrule
\multirow{4}{*}{\rotatebox{90}{Retrieval}} 
& GraftNet & 66.4 & 60.4 & 36.8 & 32.7 \\
& PullNet & 68.1 & - & 45.9 & - \\
& SR+NSM & 68.9 & 64.1 & 50.2 & 47.1 \\
& SR+NSM+E2E & 69.5 & 64.1 & 49.3 & 46.3 \\
\midrule
\multirow{6}{*}{\rotatebox{90}{LLMs}} 
& Flan-T5-xl & 31.0 & - & 14.7 & - \\
& Alpaca-7B & 51.8 & - & 27.4 & - \\
& Llama3-8B & 30.3 & 25.7 & 30.5 & 27.8 \\
& Qwen2.5-7B & 28.4 & 23.7 & 25.9 & 24.1 \\
& ChatGPT & 66.8 & - & 39.9 & - \\
& ChatGPT+CoT & 75.6 & - & 48.9 & - \\
\midrule
\multirow{13}{*}{\rotatebox{90}{\makecell[c]{LLMs\\with\\KGs}}} 
& UniKGQA & 77.2 & 72.2 & 51.2 & 49.0 \\
& KD-CoT & 68.6 & 52.5 & 55.7 & - \\
& Nutrea & 77.4 & 72.7 & 53.6 & 49.5 \\
& ToG & 81.9 & 76.0 & 68.5 & 60.2 \\
& RoG & 80.8 & 70.8 & 57.8 & 56.2 \\
& KAPING & 72.4 & 65.1 & 53.4 & 50.3 \\
& ReasoningLM & 78.5 & 71.0 & 69.0 & 64.0 \\
& FiDeLis & 84.3 & 78.3 & 71.5 & 64.3 \\
& GNN-RAG & 82.8 & 73.5 & 62.8 & 60.4 \\
& DoG & 65.4 & 55.6 & 41.0 & 46.4 \\
& DualR & 81.5 & 71.6 & 65.3 & 62.1 \\
& DP & {87.5} & {81.4} & {75.8} & {69.4} \\
& RwT & 87.0 & 79.7 & 72.4 & 66.7 \\
\midrule
& \method{} & \textbf{91.6} & \textbf{81.7} & \textbf{76.9} & \textbf{72.9} \\
\bottomrule
\end{tabular}
}
\caption{Performance comparison (\%) on WebQSP and CWQ datasets. \textbf{Bold} results indicate the best performance.}
\label{tab:main_results}

\end{table}

\paragraph{Implementation Details.}

In \method{}, GPT-4.1 serves as the backbone for context generation, candidate retrieval, relation re-ranking, and exploration summarization~\citep{liu2023evaluating}. Following prior work~\cite{sun2023think, ma2024think}, the reasoning process follows a beam search paradigm with a maximum number of iterations $I=30$ and a maximum length of $L=4$ hops. At each step, the retriever proposes $k=3$ candidate relations for the WebQSP dataset and $k=4$ for the CWQ dataset, which are then filtered and re-ranked by the Dual-Feedback Re-ranking module under a threshold of $\zeta=0.5$. Following prior work~\citep{luo2023reasoning,wang2025dynamically,ma2025deliberation}, performance is evaluated using Hits@1 and F1 score, capturing both exact-match accuracy and robustness in the presence of multiple valid answers.

\begin{table}[t]
\centering
\small
\resizebox{\linewidth}{!}{
\begin{tabular}{lcccc}
\toprule
\multirow{2}{*}{Method} & \multicolumn{2}{c}{WebQSP} & \multicolumn{2}{c}{CWQ} \\
 & Hits@1 & F1 & Hits@1 & F1 \\
\midrule
\method{} (Llama2-13B) & 88.1 & 78.3 & 74.3 & 70.7 \\
\method{} (Qwen3-14B) & 88.4 & 79.1 & 73.9 & 70.1 \\
\method{} (GPT 4.1-mini) & 90.3 & 80.6 & 75.7 & 71.3 \\
\method{} (GPT 4.1) &  \textbf{91.6} & \textbf{81.7} & \textbf{76.9} & \textbf{72.9} \\
\bottomrule
\end{tabular}
}
\caption{Performance of \method{} using different LLM-based planners as backbones on the WebQSP and CWQ datasets. \textbf{Bold} values denote the best results.}

\label{tab:llm_backbone}
\end{table}

\subsection{Performance Comparison}

Table~\ref{tab:main_results} reports the performance of \method{} against state-of-the-art baselines on KGQA datasets. The results yield the following observations: (1) Semantic parsing methods map questions into logical forms, while retrieval-based methods extract candidate subgraphs for reasoning. Although these approaches provide interpretability and capture structural semantics, semantic parsing struggles with diverse query patterns, and retrieval methods separate retrieval from reasoning, limiting adaptability in multi-hop inference. In contrast, LLMs with KGs integrate question understanding with structured graph exploration, enabling more flexible path construction and stronger generalization. (2) General-purpose LLMs, such as Flan-T5-xl, Alpaca-7B, and Llama3-8B, rely on strong language understanding and broad contextual reasoning. They can handle diverse queries and generate coherent answers, but without grounding in structured graph semantics they often produce hallucinations or logically inconsistent results, particularly in multi-hop scenarios. In contrast, LLMs with KGs explicitly leverage graph structure to constrain reasoning and ensure factual consistency. As a result, general LLMs consistently underperform compared with LLM+KG methods. (3) LLMs with KGs methods have emerged as the leading paradigm in current KGQA research, as they couple the semantic fluency of LLMs with the structural rigor of knowledge graphs. By explicitly grounding reasoning in KGs, these methods reduce hallucination, ensure factual consistency, and achieve stronger generalization to compositional and multi-hop queries. Recent advances, such as DP and RwT, further illustrate the benefits of incorporating question-specific priors and structured exploration, confirming the effectiveness of this hybrid paradigm. Compared with prior methods, \method{} delivers improved performance, attributed to two main innovations: (i) the use of contextual narratives, which preserve semantic coherence across reasoning steps and prevent logical drift, and (ii) the generalization of prior exploration trajectories into reusable experiential priors, which guide subsequent reasoning and reduce redundant exploration. Together, these components enable \method{} to outperform existing LLM+KG approaches with more coherent and robust multi-hop reasoning.

\subsection{Impact of Different LLMs}

To evaluate the impact of different LLMs as backbone within the \method{} framework, we evaluate several backbones including Llama2-13B~\citep{roque2025evolution}, Qwen3-14B~\citep{team2024qwen2}, GPT-4.1-mini, and GPT-4.1, as reported in Table~\ref{tab:llm_backbone}. The results reveal that: (1) The effectiveness of \method{} depends on the reasoning capacity of the underlying LLMs. The consistent performance gap between medium-scale open-source models and larger proprietary models illustrates the sensitivity of KGQA to planner quality, indicating that more capable LLMs enable more accurate relation selection and more stable multi-hop reasoning. (2) \method{} remains effective even with relatively weaker planners. When instantiated with Llama2-13B, the framework still outperforms strong baselines such as RwT and FiDeLis on both datasets. This indicates that although high-capacity LLMs bring additional improvements, the design of \method{} ensures robustness and competitive performance across a wide range of backbones.

\begin{table}[t]
\centering
\small
\begin{tabular}{l|cc|cc}
\toprule
\multirow{2}{*}{Method} & \multicolumn{2}{c}{WebQSP} & \multicolumn{2}{c}{CWQ} \\
 & Hits@1 & F1 & Hits@1 & F1 \\
\midrule
% 去掉context
\method{} w/o CT & 89.0 & 78.0 & 74.1 & 70.6\\

% 去掉错误信息
\method{} w/o ER & 88.6 & 77.4 & 74.8 & 71.1 \\

% 两个都去掉
\method{} w/o ALL & 87.5 & 76.7 & 73.6 & 69.9 \\
\midrule
\method{} &  \textbf{91.6} & \textbf{81.7} & \textbf{76.9} & \textbf{72.9} \\
\bottomrule
\end{tabular}
\caption{The results of ablation studies on the WebQSP and CWQ datasets. \textbf{Bold} results indicate the best performance.}

\label{tab:ablation_results}
\end{table}

\subsection{Ablation Study}

We conduct ablation studies to evaluate the contributions of key components in \method{}: (1) \method{} w/o CT, removing the dynamic context generation module; (2) \method{} w/o ER, removing the exploration generalization mechanism; and (3) \method{} w/o ALL, removing both modules simultaneously.

Experimental results are summarized in Table~\ref{tab:ablation_results}. From the results, we find that:
 (1) Removing the dynamic context generation module (\method{} w/o CT) leads to a clear performance drop, indicating that maintaining semantic coherence across reasoning steps is crucial for accurate path reasoning. Without this component, the model cannot ensure that intermediate decisions remain aligned with the overall query goal. (2) Removing the exploration generalization mechanism (\method{} w/o ER) leads to degraded performance, indicating that abstracting exploration priors into reusable knowledge is crucial for improving model performance. Without this component, the model is more likely to repeat similar exploration patterns, reducing the overall reasoning performance of \method{}. (3) When both modules are removed simultaneously (\method{} w/o ALL), the proposed \method{} suffers the largest drop in performance. This result shows that dynamic context generation and exploration generalization are complementary, and their combined effect is critical to realizing the effectiveness of \method{}.

\begin{figure*}[t]
    \centering
    \begin{subfigure}{0.325\linewidth}
        \centering
        \includegraphics[width=\linewidth]{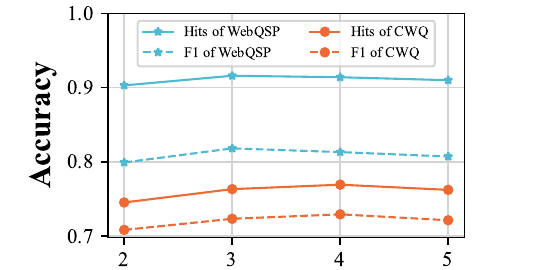}
        \caption{Number of selected relations $k$.}
        \label{fig:mcts_top_k}
    \end{subfigure}
    \hfill
    \begin{subfigure}{0.325\linewidth}
        \centering
        \includegraphics[width=\linewidth]{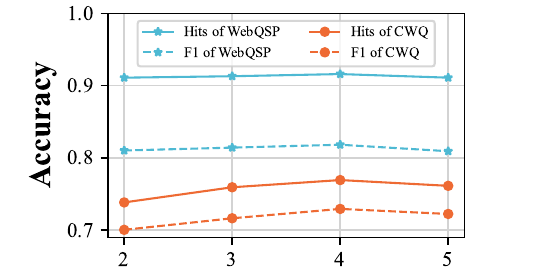}
        \caption{Reasoning path length $L$.}
        \label{fig:mcts_length}
    \end{subfigure}
    \hfill
    \begin{subfigure}{0.325\linewidth}
        \centering
        \includegraphics[width=\linewidth]{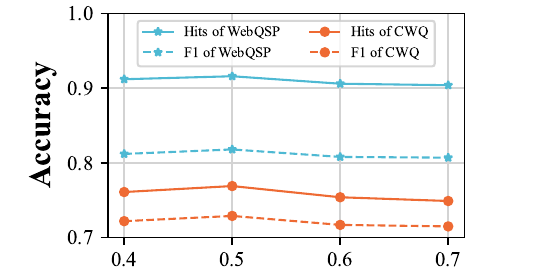}
        \caption{Threshold $\zeta$.}
        \label{fig:mcts_threshold}
    \end{subfigure}

    \caption{Sensitivity analysis of hyperparameters on the WebQSP and CWQ datasets.}

    \label{fig:hyper_sensitive}
\end{figure*}

\subsection{Sensitivity Analysis}

We perform a sensitivity analysis to examine the effect of three key hyperparameters in \method{}: the number of selected relations $k$, the maximum reasoning path length $L$, and the confidence threshold $\zeta$. The parameter $k$ specifies how many candidate relations are proposed by the LLM-based retriever at each step, $L$ determines the maximum number of hops allowed during path construction, and $\zeta$ is the threshold for relation selection in path expansion.

Figure~\ref{fig:hyper_sensitive} presents the performance of \method{} on the WebQSP and CWQ datasets as we vary $k, L$ within the range of $\{2,3,4,5\}$ and $\zeta$ within the range of $\{0.4,0.5,0.6,0.7\}$. From the results, we find that: (1) As shown in Figure~\ref{fig:hyper_sensitive}(a), increasing $k$ initially improves performance, but the gains plateau and eventually decline as the search space expands. Larger values of $k$ promote broader relational exploration but also introduce irrelevant candidates and additional cost, while smaller values restrict the diversity of relations. To balance these trade-offs, we set $k=3$ for WebQSP and $k=4$ for CWQ as the default configurations. (2) As shown in Figure~\ref{fig:hyper_sensitive}(b), extending the reasoning path length $L$ yields diminishing returns beyond a certain depth. On both WebQSP and CWQ, performance improves steadily from $L=2$ to $L=4$, but declines at $L=5$, suggesting that moderately deep paths are most effective while overly long horizons introduce noise. To balance accuracy and efficiency, we adopt $L=4$ as the default path length in all experiments. (3) As shown in Figure~\ref{fig:hyper_sensitive}(c), the confidence threshold $\zeta$ has a substantial impact on relation selection. Lower thresholds allow noisy candidates that reduce precision, whereas higher thresholds overly restrict expansion and harm recall. A moderate value provides the best trade-off, and we therefore set $\zeta=0.5$ as the default value. 

\begin{table}[t]
\centering
\small
% \vspace{-0.1cm}
\begin{tabular}{l@{\hspace{2em}}|cc|cc}
\toprule
\multirow{2}{*}{Method} & \multicolumn{2}{c|}{WebQSP} & \multicolumn{2}{c}{CWQ} \\
 & \#Tokens & \#Calls & \#Tokens & \#Calls \\
\midrule

DoG       & 22,538 & 30.9  & 37,741 & 58.1 \\
ToG       & 16,372 &  23.2 & 26,183 & 41.9 \\
RwT       & 10,680 & 15.1 & 17,885 &  28.6 \\
\midrule

\method{} & \textbf{8,782} & \textbf{14.2} & \textbf{16,414} & \textbf{27.8} \\

\bottomrule
\end{tabular}
\caption{Statistics of average number of LLM calls and token consumption per question on WebQSP and CWQ datasets.}

\label{tab:Efficiency}
\end{table}

\begin{table*}[t]
\renewcommand{\arraystretch}{1.3}
\small
\centering
% \vspace{-0.1cm}
\begin{tabular}{>{\centering\arraybackslash}m{2.5cm}|p{12.2cm}}
\toprule
\textbf{Question} & What featured Kim Richards and was producted by Ann Marie Sanerlin? \\
\hline
\textbf{Answer} & \textcolor{red}{Race to Witch Mountain} \\
\hline
\multirow{2}{*}{\textbf{Llama-2-13B}} & The TV movie \textcolor{blue}{"The Christmas Card" (2006)} featured Kim Richards and was produced by Anne Marie Sanerlin, however I couldn't verify if it matches exactly "Ann Marie Sanerlin". \\
\hline
\multirow{3}{*}{\textbf{Qwen-3-14B}} & The 2019 film \textcolor{blue}{"The Last Days of American Crime"} featured Kim Richards and was produced by Ann Marie Sanerlin. However, this information is not widely documented, and Sanerlin's role in the production is not prominently noted in mainstream sources. \\
\hline
\multirow{3}{*}{\textbf{GPT 4.1}} & The project that featured Kim Richards and was produced by Ann Marie Sanerlin is the 1977 television film \textcolor{blue}{"The Car."} Kim Richards starred in the film, and Ann Marie Sanerlin was credited as the producer. \\
\hline
\multirow{3}{*}{\textbf{GPT 4.1-mini}} & The project featuring Kim Richards and produced by Ann Marie Sanerlin is the reality TV series \textcolor{blue}{"The Real Housewives of Beverly Hills."} Kim Richards is a cast member, and Ann Marie Sanerlin has served as a producer on the show. \\
\hline
\multirow{11}{*}{\textbf{\method{}}} & \texttt{\textbf{Reasoning Path}}: Entity (id: 1384221) $\rightarrow$ film.actor.film $\rightarrow$
film.performance.film $\rightarrow$
\textcolor{red}{Race to Witch Mountain}. \\
& \texttt{\textbf{Contextual Narrative}}: Identify the films featuring Kim Richards by tracing her acting roles and performances, and then determine which of these films were produced by Ann Marie Sanerlin. \\
& \texttt{\textbf{Exploration Paths}}: (i) tv.tv\_actor.starring\_roles $\rightarrow$ tv.regular\_tv\_appearance.series $\rightarrow$ tv.tv\_program.regular\_cast $\rightarrow$ tv.regular\_tv\_appearance.series; (ii) tv.tv\_actor.guest\_roles $\rightarrow$ tv.tv\_guest\_role.episodes\_appeared\_in; (iii) tv.tv\_personality.tv\_regular\_appearances $\rightarrow$ tv.tv\_regular\_personal\_appearance.program $\rightarrow$ tv.tv\_program.regular\_personal\_appearances $\rightarrow$ tv.tv\_regular\_personal\_appearance.program. \\
& \texttt{\textbf{Exploration Patterns}}: Repeatedly traversing between acting roles, appearances, and program or episode nodes without a direct connection to the target individual (Ann Marie Sanerlin) leads to unproductive loops or dead ends. Over-reliance on cast or appearance relationships, especially when the target is a producer rather than an actor, tends to exceed path limits without yielding relevant results. Future reasoning should prioritize direct production or crew-related links over cast or appearance-based paths when seeking information about behind-the-scenes personnel. \\
\bottomrule
\end{tabular}
\caption{Case study of \method{}. We highlight the correct answers in \textcolor{red}{Red} and the wrong answers in \textcolor{blue}{Blue}.}

\label{tab:case_study1}
\end{table*}

\subsection{Computation Consumption Analysis}

Table~\ref{tab:Efficiency} compares the computation consumption of different KGQA methods in terms of LLM calls and token usage, where \method{} achieves lower token consumption and fewer LLM calls. On the WebQSP dataset, \method{} lowers token usage from 10,680 to 8,782 and LLM calls from 15.1 to 14.2, outperforming RwT. On the CWQ dataset, it further reduces tokens from 17,885 to 16,414 and calls from 28.6 to 27.8, again demonstrating clear efficiency gains over RwT. These efficiency gains arise from \method{}, in which contextual narratives preserve semantic coherence across reasoning steps while exploration generalization reduces redundant exploration. As a result, \method{} lowers both LLM call frequency and token consumption, yielding a more efficient and scalable reasoning framework than existing baselines. More results on worst or failure cases can be found in Appendix.~\ref{sec:computation}.

\subsection{Case study}

Table~\ref{tab:case_study1} presents a case study comparing the outputs of \method{} with four representative LLMs: Llama-2-13B, Qwen-3-14B, GPT 4.1, and GPT 4.1-mini. For the query \textit{“What featured Kim Richards and was produced by Ann Marie Sanerlin?”}, all baseline models produce incorrect answers, often hallucinating films or conflating unrelated television productions. In contrast, \method{} correctly identifies \textit{Race to Witch Mountain} by following a structured \texttt{Reasoning Path} that connects Kim Richards to her film performances and integrates production information to refine the candidate set. The reasoning process of \method{} begins by locating entities associated with Kim Richards and expanding along film-related relations that represent her acting roles. At each hop, the \texttt{Contextual Narrative} guides the model to remain semantically aligned with the query, first retrieving films featuring Kim Richards and then filtering them according to the production information linked to Ann Marie Sanerlin. This narrative grounding ensures that relation selection remains coherent and context-aware throughout the multi-hop reasoning process. In parallel, \texttt{Exploration Patterns} abstracted from prior exploration trajectories capture recurring unproductive behaviors, such as cast–series loops that fail to satisfy producer-related constraints. Through the interplay of narrative guidance and exploration prior, \method{} effectively narrows the search space and converges on the correct answer. Additional case study is provided in Appendix~\ref{sec:case_study}.

\section{Conclusion}
\vspace{-0.3cm}
In this paper, we propose \textbf{T}rajectory-aware \textbf{R}easoning with \textbf{A}daptive \textbf{C}ontext and \textbf{E}xploration priors (\method{}), an experiential framework that enhances the coherence and robustness of multi-hop knowledge graph question answering (KGQA). \method{} comprises three synergistic components: dynamic context generation, which preserves semantic continuity across reasoning steps; exploration generalization, which abstracts prior exploration trajectories into reusable experiential priors; and dual-feedback re-ranking, which integrates contextual and experiential signals to guide relation selection. Extensive experiments on the WebQSP and CWQ datasets demonstrate that \method{} consistently outperforms state-of-the-art baselines.
\section*{Limitations}

Although \method{} demonstrates strong empirical performance, it still has several limitations. The framework currently relies on manually designed prompts and exploration priors, which may limit its adaptability to heterogeneous or noisy knowledge graphs. Moreover, despite improved efficiency compared to prior approaches, the integration of multiple reasoning components introduces non-trivial inference costs. Future work will explore automatic prompt optimization, more lightweight retrieval mechanisms, and reinforcement learning–based controllers to further improve scalability and generalization. Extending \method{} beyond KGQA to domains such as scientific discovery and multi-agent collaboration also represents a promising direction.

\bibliography{references}

\newpage
\appendix
\onecolumn
\section{Appendix}

\subsection{Algorithm}\label{sec:algorithm}

\begin{algorithm}[h]
    \caption{Trajectory-aware Reasoning with Adaptive Context and Exploration Priors (\method{})}
    \renewcommand{\algorithmicrequire}{\textbf{Input:}}
    \renewcommand{\algorithmicensure}{\textbf{Output:}}
    \begin{algorithmic}[1]
        \REQUIRE Question $q$, knowledge graph $\mathcal{K}$, 
        maximum reasoning steps $L$, \\ candidate pool size $k$, confidence threshold $\zeta$, 
        maximum iterations $I$
        \ENSURE Predicted answer entity set $\mathcal{A}$

        \STATE Initialize exploration priors $\mathcal{F} \leftarrow \emptyset$
        \STATE Initialize current reasoning path $\mathcal{P}_q^{(0)}$ from topic entities $\mathcal{E}_q$
        
        \FOR{$i = 1$ to $I$}
            \FOR{$t = 1$ to $L$}
                \STATE \textbf{Dynamic Context Generation:}  
                Generate contextual narrative  
                $C_q^{(t)} = f_{\text{ctx}}(q, R_q^{(t-1)})$

                \STATE \textbf{Candidate Retrieval:}  
                Retrieve top-$k$ candidate relations from the local neighborhood  
                $\mathcal{C}_t = f_{\text{ret}}(q, C_q^{(t)}, \mathcal{N}(e_t))$

                \STATE \textbf{Dual-Feedback Re-ranking:}  
                For each $r \in \mathcal{C}_t$, compute relevance score  
                $s(r) = f_{\text{rank}}(q, R_q^{(t)}, r, \mathcal{F})$,  
                and expand the reasoning path with relations satisfying $s(r) \geq \zeta$
            \ENDFOR

            \IF{$\mathcal{P}_q^{(t)} \in \mathcal{T}_{\text{depth}} \cup \mathcal{T}_{\text{expand}}$}
                \STATE \textbf{Trajectory Summarization:}  
                Generate exploration summary  
                $D_q = f_{\text{sum}}(q, R_q^{(t)})$

                \STATE \textbf{Exploration Generalization:}  
                Update exploration priors  
                $\mathcal{F} \leftarrow f_{\text{gen}}(\mathcal{F} \cup \{D_q\})$
            \ENDIF
        \ENDFOR

        \STATE \textbf{return} $\mathcal{A}$ from the reasoning trajectory with the highest relevance score
    \end{algorithmic}
\end{algorithm}

\subsection{Datasets}\label{sec:dataset}

\subsubsection{Dataset Description}

To evaluate the effectiveness of the proposed \method{}, we conduct extensive experiments on two widely adopted multi-hop Knowledge Graph Question Answering (KGQA) benchmarks: WebQSP~\cite{talmor2018web} and CWQ~\cite{yih2016value}. The detailed description of each dataset is provided below:

\begin{itemize}
    \item WebQSP is a widely adopted benchmark for multi-hop KGQA, derived from the WebQuestions dataset~\cite{yih2016value}. Each query is paired with an annotated SPARQL program over Freebase, enabling precise evaluation of reasoning accuracy. The benchmark emphasizes complex reasoning phenomena such as multi-hop inference, relation composition, and entity disambiguation, making it a rigorous testbed for KGQA models. It comprises 2,848 training, 250 validation, and 1,639 test instances.
    \item ComplexWebQuestions (CWQ) is a large-scale benchmark for evaluating compositional reasoning in KGQA. It is derived by decomposing seed queries from WebQuestionsSP into multiple sub-questions and recombining them into more complex forms~\cite{talmor2018web}. Each question is annotated with SPARQL programs over Freebase, enabling systematic assessment of multi-hop reasoning. Compared with WebQSP, CWQ introduces longer reasoning chains and richer compositional structures, making it a more challenging benchmark. The dataset contains 27,639 training, 3,519 validation, and 3,531 test instances.
\end{itemize}

\subsubsection{Data processing}

Following prior work~\cite{shen2025reasoning,wang2025dynamically}, we preprocess these two benchmarks by constructing localized subgraphs centered on the topic entity of each question to reduce the search space. For every query in WebQSP~\cite{yih2016value} and CWQ~\cite{talmor2018web}, we extract a subgraph from Freebase that includes all triples within a predefined number of hops from the question entity. This strategy retains the essential contextual information needed for multi-hop reasoning while substantially improving computational efficiency. Following prior work~\cite{sun2023think,wang2025dynamically}, we randomly sample 1,000 questions from the test set of each dataset for evaluation. 

\subsection{ Baselines}\label{sec:baseline}

In this part, we introduce the details of the compared baselines as follows:

\begin{itemize}

\item \textbf{Semantic Parsing Methods.} We compare our \method{} with seven semantic parsing methods:

\begin{itemize}
    \item \textbf{KV-Mem}: KV-Mem~\cite{miller2016key} facilitate direct document reading by encoding information into key–value memory slots, enabling efficient retrieval and integration of relevant content for question answering without relying solely on sequential context.
    \item \textbf{EmbedKGQA}: EmbedKGQA~\cite{saxena2020improving} enhances multi-hop question answering over knowledge graphs by leveraging pretrained knowledge base embeddings, enabling effective reasoning over entities and relations without the need for explicit path enumeration.
    \item \textbf{QGG}: QGG~\cite{lan2020query} tackles multi-hop complex question answering over knowledge bases by constructing query graphs that capture the compositional structure of questions, enabling systematic reasoning across entities and relations.
    \item \textbf{NSM}: NSM~\cite{he2021improving} advances multi-hop knowledge base question answering through a teacher–student framework, where the teacher network employs bidirectional reasoning to produce intermediate supervision signals, enabling the student network to learn more reliable reasoning paths and reduce spurious inference.
    \item \textbf{TransferNet}: TransferNet~\cite{shi2021transfernet} provides an effective and transparent framework for multi-hop question answering over relation graphs by transferring relational evidence across hops, thereby enhancing interpretability and reasoning accuracy.
    \item \textbf{KGT5}: KGT5~\cite{saxena2022sequence} addresses knowledge graph completion and question answering within a unified sequence-to-sequence framework, enabling the model to jointly generate missing triples and answer queries through end-to-end training.
    \item \textbf{DECAF}: DECAF~\cite{yu2022decaf} jointly decodes both answers and logical forms for question answering over knowledge bases, enabling the model to generate interpretable reasoning paths alongside answers within a unified sequence generation framework.
    
\end{itemize}

\item \textbf{Retrieval-Based Methods.} We compare our \method{} with four retrieval-based methods:

\begin{itemize}
    \item \textbf{GraftNet}: GraftNet~\cite{sun2018open} addresses open-domain question answering by integrating knowledge bases and textual evidence in an early fusion framework, allowing the model to jointly reason over structured and unstructured information sources.
    \item \textbf{PullNet}: PullNet~\cite{sun2019pullnet} enables open-domain question answering by iteratively retrieving relevant facts from both knowledge bases and text, dynamically constructing a subgraph to support multi-hop reasoning across heterogeneous information sources.
    \item \textbf{SR+NSM}: SR+NSM~\cite{zhang2022subgraph} enhances multi-hop knowledge base question answering by integrating a subgraph retrieval mechanism with a neural state machine, enabling the model to first retrieve a relevant subgraph and then perform stepwise reasoning within this focused context for improved answer accuracy.
    \item \textbf{SR+NSM+E2E}: SR+NSM+E2E~\cite{zhang2022subgraph} further advances multi-hop knowledge base question answering by jointly training subgraph retrieval and neural state machine reasoning in an end-to-end manner, enabling more effective integration between subgraph selection and multi-hop inference for improved answer prediction.
\end{itemize}

    \item  \textbf{General
Large Language Models (LLMs).} We compare our \method{} with six general
LLMs:

\begin{itemize}

    \item \textbf{Flan-T5-xl}: Flan-T5-xl~\cite{chung2024scaling} is a large-scale instruction-finetuned language model that builds on the T5 architecture, achieving superior generalization across diverse tasks by leveraging fine-tuning on a broad set of instructional prompts.
    \item \textbf{Alpaca-7B}: Alpaca-7B~\cite{taori2023stanford} is an instruction-following language model based on LLaMA, developed by Stanford, and fine-tuned on a curated set of instructional demonstrations to enhance generalization and task-following abilities.
    \item \textbf{Llama3-8B}: Llama3-8B~\cite{dubey2024llama} is a next-generation instruction-following language model from Meta, featuring 8 billion parameters and trained with advanced alignment techniques to deliver robust performance across a wide range of tasks.
    \item \textbf{Qwen2.5-7B}: Qwen2.5-7B~\cite{team2024qwen2} is a 7-billion-parameter instruction-tuned language model from Alibaba, designed for high-quality, general-purpose reasoning and enhanced task following through extensive multi-domain instruction fine-tuning.
    \item \textbf{ChatGPT}: ChatGPT~\cite{schulman2022chatgpt} is a conversational large language model developed by OpenAI, optimized for dialogue and task-oriented interactions through extensive instruction tuning and reinforcement learning from human feedback.
    \item \textbf{ChatGPT+CoT}: ChatGPT with Chain-of-Thought (CoT)~\cite{wei2022chain} augments the standard ChatGPT model with step-by-step reasoning capabilities, enabling more interpretable and accurate responses on complex tasks through explicit multi-step thought processes.
\end{itemize}

    \item \textbf{LLMs with KG.} We compare our \method{} with thirteen LLMs with KG methods:

\begin{itemize}
    \item  \textbf{UniKGQA}: UniKGQA~\cite{jiang2022unikgqa} unifies retrieval and reasoning for multi-hop question answering over knowledge graphs, jointly optimizing entity retrieval and reasoning steps within an end-to-end framework to enhance answer accuracy and reasoning transparency.
    \item \textbf{KD-CoT}: KD-CoT~\cite{wang2023knowledge} explores faithful reasoning in large language models for knowledge-intensive question answering by integrating knowledge-grounded chain-of-thought prompting, thereby improving factual accuracy and interpretability in multi-hop reasoning tasks.
    \item \textbf{Nutrea}: Nutrea~\cite{choi2023nutrea} introduces a neural tree search framework for context-guided multi-hop knowledge graph question answering, enabling efficient exploration and aggregation of reasoning paths guided by both question context and graph structure.
    \item \textbf{ToG}: ToG~\cite{sun2023think} is a framework that enables large language models to perform deep and responsible reasoning over knowledge graphs by combining structured graph information with iterative thinking and verification mechanisms for reliable multi-hop QA.
    \item \textbf{RoG}: RoG~\cite{luo2023reasoning} enables deep and responsible reasoning of large language models over knowledge graphs by explicitly modeling structured multi-hop inference, enhancing both reasoning transparency and factual reliability in complex question answering scenarios.
    \item \textbf{KAPING}: KAPING~\cite{baek2023knowledge} promotes faithful and interpretable large language model reasoning by leveraging explicit graph-based structures, ensuring transparent multi-hop inference and improved answer reliability for knowledge-intensive tasks.
    \item \textbf{ReasoningLM}: ReasoningLM~\cite{jiang2023reasoninglm} enables structural subgraph reasoning in pre-trained language models for question answering over knowledge graphs, allowing the model to explicitly exploit subgraph structures for more accurate and interpretable multi-hop reasoning.
    \item \textbf{FiDeLis}: FiDeLis~\cite{sui2024fidelis} enables faithful reasoning in large language models for knowledge graph question answering by integrating explicit logical constraints and stepwise supervision, thereby improving factual consistency and interpretability in multi-hop inference.

    \item \textbf{GNN-RAG}: GNN-RAG~\cite{mavromatis2024gnn} enhances large language model reasoning by employing graph neural retrieval, enabling the model to incorporate structured knowledge from graphs for more accurate and context-aware question answering.
    
    \item \textbf{DoG}: DoG~\cite{ma2025debate} presents a flexible and reliable reasoning framework for large language models, leveraging debate-style interactions over knowledge graphs to improve reasoning transparency, flexibility, and answer robustness.
   
    \item \textbf{DualR}: DualR ~\cite{liu2025dual} introduces a collaborative framework for knowledge graph question answering, where graph neural networks and large language models jointly perform complementary reasoning to enhance multi-hop inference accuracy and interpretability.
    \item \textbf{DP}: DP~\cite{ma2025deliberation} facilitates trustworthy reasoning of large language models on knowledge graphs by incorporating prior knowledge and iterative deliberation, thereby enhancing reliability, transparency, and robustness in complex question answering tasks.
    
    \item \textbf{RwT}: RwT~\cite{shen2025reasoning} achieves faithful question answering over knowledge graphs by organizing multi-hop inference as a tree-based reasoning process, ensuring interpretable and reliable answer generation.

\end{itemize}

\end{itemize}

\begin{table}[t]
\centering
\small
\begin{tabular}{l@{\hspace{2em}}|cc|cc}
\toprule
\multirow{2}{*}{Setting} & \multicolumn{2}{c|}{WebQSP} & \multicolumn{2}{c}{CWQ} \\
 & \#Tokens & \#Calls & \#Tokens & \#Calls \\
\midrule

Worst   & 13,787 & 22.3 & 29,381 & 46.3 \\
Failure & 10,623 & 16.3 & 20,849 & 34.0 \\
Average & \textbf{8,782} & \textbf{14.2} & \textbf{16,414} & \textbf{27.8} \\
\bottomrule
\end{tabular}
\caption{Statistics of average number of LLM calls and token consumption per question under worst and failure cases on the WebQSP and CWQ datasets 
datasets.}
\label{tab:efficiency_new}
\end{table}

\subsection{More Computation Consumption Analysis}\label{sec:computation}

To provide a more comprehensive scalability analysis, we re-ran the experiments on WebQSP and CWQ and recorded per-question token consumption and LLM call statistics. The corresponding worst-case and failure-case results are reported in Table~\ref{tab:efficiency_new}. Notably, in failure mode, both token usage and LLM calls are only moderately higher than the average. As described in Section~\ref{sec:eg}, reasoning expansion terminates when the LLM determines that no valid relations can be proposed. Although failure cases incur slightly higher costs than the average due to difficulty, the mechanism ensures they terminate essentially when no valid expansions are found, preventing the exponential cost blow-up typically seen in unconstrained search. In contrast, worst-case instances involve deeper exploration across multiple candidate paths, resulting in higher computational cost. Even in this scenario, token usage increases to approximately 1.5× the average on the WebQSP dataset and 1.7× on the CWQ dataset, demonstrating bounded growth rather than uncontrolled expansion. Furthermore, the exploration prior mechanism suppresses redundant or previously identified unproductive exploration patterns, preventing repeated branching and mitigating the risk of exponential cost escalation.

\subsection{More case studies}\label{sec:case_study}

Table~\ref{tab:case_study2} presents a case study comparing the outputs of \method{} with four representative LLMs: Llama-2-13B, Qwen-3-14B, GPT 4.1, and GPT 4.1-mini. For the query \textit{What guitar does Corey Taylor play?}, all baseline models produce incorrect answers, often attributing instruments such as Gibson, Jackson, or Fender guitars. In contrast, \method{} correctly identifies \textit{Bass guitar} by traversing a structured \texttt{Reasoning Path} that links Corey Taylor to his instrument associations in the knowledge graph. This process is guided by a \texttt{Contextual Narrative} that remains semantically aligned with the question, allowing the model to prioritize instrument-related relations over indirect contribution-based paths. In parallel, \texttt{Exploration Patterns} capture recurring unproductive traversal behaviors, guiding the search away from contribution- or recording-centric relations that fail to surface instrument information. As illustrated in this case, \method{} efficiently arrives at the correct answer while preserving interpretability through explicit KG-grounded reasoning.

\subsection{Prompts}

To elucidate the role of prompting in \method{}, we highlight the prompt designs within its three core components: (i) In Dynamic Context Generation, prompts are designed to transform the evolving relation sequence into coherent contextual narratives, thereby ensuring that each reasoning step remains consistent with both the original query and the accumulated path, as shown in Figure.~\ref{fig:prompt_context_generation}. (ii) In Exploration Generalization,
prompts guide the LLM to summarize terminated exploration trajectories and distill recurring exploration patterns into reusable experiential priors, as illustrated in Figure~\ref{fig:prompt_trajectory_summary} and Figure~\ref{fig:prompt_exploration_patterns}. (iii) In Dual-Feedback Re-ranking, prompts are employed in a two-stage process: the first stage retrieves a top-$k$ set of candidate relations conditioned on the contextual narrative, and the second stage re-ranks these candidates by jointly considering the evolving relation sequence and the exploration priors, as shown in Figure.~\ref{fig:prompt_candidate_retrieval} and ~\ref{fig:prompt_reranking}.

\newpage

\begin{table*}[h]
\renewcommand{\arraystretch}{1.3}
\small
\centering
% \vspace{-0.1cm}
\begin{tabular}{>{\centering\arraybackslash}m{2.5cm}|p{12.2cm}}
\toprule
\textbf{Question} &What guitar does Corey Taylor play? \\
\hline
\textbf{Answer} & \textcolor{red}{Bass guitar} \\
\hline
\multirow{3}{*}{\textbf{Llama-2-13B}} & Corey Taylor, the lead vocalist of Slipknot and Stone Sour, is often seen playing various guitars, but his primary guitar is a custom Ibanez RG, and he also uses other Ibanez models and \textcolor{blue}{Dean Guitars}. \\
\hline
\multirow{3}{*}{\textbf{Qwen-3-14B}} & Corey Taylor, known for his work with Slipknot and Stone Sour, primarily plays \textcolor{blue}{Jackson JS Series guitars}, including the JS69 and JS32 models. He also uses his signature Jackson Corey Taylor JS guitars, designed for heavy rock and metal styles. \\
\hline
\multirow{3}{*}{\textbf{GPT 4.1}} & Corey Taylor, primarily known as a vocalist for Slipknot and Stone Sour, occasionally plays guitar. He has been seen using Gibson Explorer, Gibson Les Paul, and \textcolor{blue}{Fender Telecaster guitars} during live performances and studio sessions, especially with Stone Sour. \\
\hline
\multirow{3}{*}{\textbf{GPT 4.1-mini}} & Corey Taylor, lead vocalist of Slipknot and Stone Sour, primarily plays \textcolor{blue}{Gibson guitars}, especially the Gibson Les Paul models. He has also been seen using Fender Stratocasters and other custom guitars during performances. His Gibson Les Paul Custom is one of his signature instruments. \\
\hline
\multirow{15}{*}{\textbf{\method{}}} & \texttt{\textbf{Reasoning Path}}: Entity (id: 1242859) $\rightarrow$ music.group\_member.instruments\_played $\rightarrow$
\textcolor{red}{Bass guitar}. \\
& \texttt{\textbf{Contextual Narrative}}: Identify Corey Taylor's musical contributions, determine the recordings he contributed to, and then find the tracks associated with those contributions to discover what guitar he plays.\\
& \texttt{\textbf{Exploration Paths}}: (i) music.artist.contribution $\rightarrow$ music.recording\_contribution.contributor $\rightarrow$  music.artist.track\_contributions $\rightarrow$  music.track\_contribution.role; (ii)  music.artist.contribution $\rightarrow$ music.recording\_contribution.contributor $\rightarrow$ music.artist.contribution $\rightarrow$ music.recording\_contribution.album; (iii) music.artist.contribution $\rightarrow$ music.recording\_contribution.contributor $\rightarrow$ music.artist.track\_contributions $\rightarrow$ music.track\_contribution.track;

 \\
& \texttt{\textbf{Exploration Patterns}}: Tracing an artist’s instrument details by following their musical contributions, recording credits, or associated tracks/albums often leads to overly long and unproductive paths that do not yield specific information about the instruments they play. These approaches tend to exhaust path depth limits without success because contribution and recording relations rarely encode direct instrument details. Future reasoning should prioritize direct biographical or equipment-related relations over indirect contribution-based paths when seeking information about an artist’s instruments. \\
\bottomrule
\end{tabular}
\caption{Case study of \method{}. We highlight the correct answers in \textcolor{red}{Red} and the wrong answers in \textcolor{blue}{Blue}.}
\label{tab:case_study2}
\end{table*}

\newpage

\begin{tcolorbox}[title=Prompt Template for Dynamic Context Generation, colback=white, colframe=black!75, boxrule=0.5pt, arc=2mm]

\textbf{Role:} \\
You are an expert assistant for Knowledge Graph Question Answering (KGQA). Your capability lies in transforming structured relation sequences into coherent natural language narratives that capture their semantics in the context of the input question.

\textbf{Task:} \\
Given a natural language question and a sequence of relations representing a reasoning path, generate a concise and contextually faithful narrative that describes the meaning of this path. The narrative will serve as dynamic context to guide subsequent reasoning steps.

\textbf{Rules and Constraints:}
\begin{itemize}[leftmargin=1.5em, itemsep=0pt, topsep=0pt]
    \item \textbf{Faithful Representation}: The narrative must accurately reflect the semantics of the provided relations without introducing external knowledge.
    \item \textbf{Conciseness}: Express the reasoning path in a clear and compact natural language form.
    \item \textbf{Context Awareness}: Ensure that the generated narrative maintains coherence with the original question and the evolving reasoning trajectory.
\end{itemize}

\textbf{Example:}
\begin{itemize}[leftmargin=1.5em, itemsep=0pt, topsep=2pt]
    \item \textbf{Input:}
    \begin{itemize}
        \item \texttt{Question}: "Where was the director of the movie Titanic born?"
        \item \texttt{Path Relations}: \{"movie.directed\_by", "person.place\_of\_birth"\}
    \end{itemize}
    \item \textbf{Output:}
\begin{verbatim}
"Find the director of the movie Titanic, and then find the birthplace 
of that person."
\end{verbatim}

    \item \textbf{Input:}
    \begin{itemize}
        \item \texttt{Question}: "What kind of currency does Germany use?"
        \item \texttt{Path Relations}: \{"country.currency\_used"\}
    \end{itemize}
    \item \textbf{Output:}
\begin{verbatim}
"Find the currency used by the country Germany."
\end{verbatim}
\end{itemize}

\textbf{Your Task}
\begin{itemize}[leftmargin=1.5em, itemsep=0pt, topsep=2pt]
    \item \texttt{Question}: \{question\}
    \item \texttt{Path Relations}: \{relations\_list\}
\end{itemize}

\textbf{Output:}
\end{tcolorbox}
\captionof{figure}{Prompt template used in the Dynamic Context Generation module to transform relation sequences into natural language narratives.}
\label{fig:prompt_context_generation}

\begin{tcolorbox}[title=Prompt Template for Trajectory Summary, colback=white, colframe=black!75, boxrule=0.5pt, arc=2mm]

\textbf{Role:} \\
You are a \textbf{“Trajectory Analyst”} responsible for analyzing terminated reasoning trajectories in Knowledge Graph Question Answering (KGQA).

\textbf{Task:} \\
Given a natural language question, an explored reasoning trajectory, and the reason why this trajectory terminated, generate a concise natural language summary that explains:
(i) what specific reasoning direction the trajectory attempted, and
(ii) why it reached a stopping point (e.g., dead end, repetitive loop, or semantic drift).
This summary should describe the specific instance and \textbf{must not} be generalized into a reusable rule.

\textbf{Rules and Constraints:}
\begin{itemize}[leftmargin=1.5em, itemsep=0pt, topsep=0pt]
    \item \textbf{Specificity}: Clearly describe the concrete reasoning direction taken by the trajectory.
    \item \textbf{Termination Awareness}: Explicitly explain why the trajectory stopped (e.g., max depth, irrelevant relations, or lack of feasible expansion).
    \item \textbf{Conciseness}: Keep the summary short and focused on the essential reasoning behavior.
    \item \textbf{No Generalization}: Do not abstract the summary into a general rule or pattern.
\end{itemize}

\textbf{Example:}
\begin{itemize}[leftmargin=1.5em, itemsep=0pt, topsep=2pt]
    \item \textbf{Input:}
    \begin{itemize}
        \item \texttt{Question}: ``Where was the director of Titanic born?''
        \item \texttt{Explored Trajectory}: \{"movie.directed\_by", ``person.spouse''\}
        \item \texttt{Reason for Termination}: Max depth reached
    \end{itemize}
    \item \textbf{Output:}
\begin{verbatim}
"The trajectory attempted to follow the director’s spouse, 
but this direction drifted away from the original question 
about the director’s birthplace."
\end{verbatim}
\end{itemize}

\textbf{Your Task}
\begin{itemize}[leftmargin=1.5em, itemsep=0pt, topsep=2pt]
    \item \texttt{Question}: \{question\}
    \item \texttt{Explored Trajectory}: \{explored\_path\}
    \item \texttt{Reason for Termination}: \{reason\_for\_termination\}
\end{itemize}

\textbf{Output:}
\end{tcolorbox}
\captionof{figure}{Prompt template used in the Exploration Generalization module to summarize terminated reasoning trajectories into trajectory-level descriptions.}
\label{fig:prompt_trajectory_summary}

\begin{tcolorbox}[title=Prompt Template for Exploration Pattern Extraction, colback=white, colframe=black!75, boxrule=0.5pt, arc=2mm]

\textbf{Role:} \\
You are a \textbf{“Pattern Recognizer”} responsible for identifying recurring patterns in reasoning trajectories for Knowledge Graph Question Answering (KGQA).

\textbf{Task:} \\
Given a list of \textbf{Trajectory Summaries} derived from recent terminated or stalled reasoning attempts, identify common reasoning patterns, shared unproductive behaviors, or structural pitfalls. Synthesize these observations into concise \textbf{Exploration Patterns} that can guide future reasoning away from similar mistakes.

\textbf{Rules and Constraints:}
\begin{itemize}[leftmargin=1.5em, itemsep=0pt, topsep=0pt]
    \item \textbf{Pattern Abstraction}: Identify recurring trends or shared characteristics across multiple trajectory summaries.
    \item \textbf{Conciseness}: Express the extracted exploration patterns in a brief, compact paragraph.
    \item \textbf{Actionability}: Phrase the patterns as guidance that can warn or steer future reasoning steps.
    \item \textbf{No Case-Specific Details}: Avoid mentioning entity names or instance-specific information.
\end{itemize}

\textbf{Example}
\begin{itemize}[leftmargin=1.5em, itemsep=0pt, topsep=2pt]
    \item \textbf{Input:}
    \begin{itemize}
        \item \texttt{Trajectory Summaries}:
        \begin{itemize}
            \item ``The trajectory explored the \texttt{spouse} relation but found no link to birthplace.''
            \item ``The trajectory followed \texttt{children}, leading to a dead end for location-related queries.''
            \item ``The trajectory checked \texttt{sibling}, which was irrelevant to the target information.''
        \end{itemize}
    \end{itemize}
    \item \textbf{Output:}
\begin{verbatim}
"Family-related relations such as spouse, children, and sibling
often lead to unproductive paths for location-focused questions.
Future reasoning should prioritize direct location or biography
relations instead."
\end{verbatim}
\end{itemize}

\textbf{Your Task}
\begin{itemize}[leftmargin=1.5em, itemsep=0pt, topsep=2pt]
    \item \texttt{Trajectory Summaries}: \{trajectory\_summaries\}
\end{itemize}

\textbf{Output:}
\end{tcolorbox}
\captionof{figure}{Prompt template used in the Exploration Generalization module to distill trajectory summaries into reusable exploration patterns.}
\label{fig:prompt_exploration_patterns}

\begin{tcolorbox}[title=Prompt Template for Candidate Retrieval (Dual-Feedback Stage 1), colback=white, colframe=black!75, boxrule=0.5pt, arc=2mm]

\textbf{Role:} \\
You are a \textbf{“Relation Retriever”} for Knowledge Graph Question Answering (KGQA).

\textbf{Task:} \\
Given a natural language question, the current reasoning context, and a list of candidate relations, select up to \texttt{k} relations that are most likely to extend the reasoning path toward the correct answer.

\textbf{Rules and Constraints:}
\begin{itemize}[leftmargin=1.5em, itemsep=0pt, topsep=0pt]
    \item \textbf{Fidelity to Candidates}: Selections must come strictly from the provided \texttt{Candidate Relations} list.
    \item \textbf{Quantity Limit}: Return no more than \texttt{k} relations. If fewer are relevant, return only those.
    \item \textbf{Output Format}: The response must be a Python-parseable list of strings. If no relation is relevant, return an empty list \texttt{[]}.
\end{itemize}

\textbf{Example}
\begin{itemize}[leftmargin=1.5em, itemsep=0pt, topsep=2pt]
    \item \textbf{Input:}
    \begin{itemize}
        \item \texttt{Question}: "Who is the CEO of Tesla?"
        \item \texttt{Current Context}: "This is the start of the path."
        \item \texttt{Candidate Relations}: \{"organization.leadership", "organization.founders", "organization.headquarters", "organization.industry"\}
        \item \texttt{K}: 2
    \end{itemize}
    \item \textbf{Output:}
\begin{verbatim}
["organization.leadership", "organization.founders"]
\end{verbatim}
\end{itemize}

\textbf{Your Task}
\begin{itemize}[leftmargin=1.5em, itemsep=0pt, topsep=2pt]
    \item \texttt{Question}: \{question\}
    \item \texttt{Current Context}: \{context\_narrative\}
    \item \texttt{Candidate Relations}: \{candidate\_relations\}
    \item \texttt{K}: \{k\}
\end{itemize}

\textbf{Output:}
\end{tcolorbox}
\captionof{figure}{Prompt template used in the Candidate Retrieval stage of Dual-Feedback Re-ranking to select top-$k$ relations.}
\label{fig:prompt_candidate_retrieval}

\begin{tcolorbox}[
  title=Prompt Template for Dual-Feedback Re-ranking (Stage 2),
  colback=white,
  colframe=black!75,
  boxrule=0.5pt,
  arc=2mm
]

\textbf{Role:} \\
You are a \textbf{``Path Evaluator''} that incorporates contextual coherence and prior exploration experience when ranking candidate relations.

\textbf{Task:} \\
Given a natural language question, a historical reasoning path, a list of top-\texttt{k} candidate relations, and a summary of past exploration experiences, evaluate the plausibility of each candidate relation as the next reasoning step. Assign a numerical score between 0.0 (bad fit) and 1.0 (perfect fit), reflecting both logical relevance to the question and consistency with effective exploration patterns.

\textbf{Rules and Constraints:}
\begin{itemize}[leftmargin=1.5em, itemsep=0pt, topsep=0pt]
  \item \textbf{Contextual Coherence}: Evaluate how well each candidate relation extends the current reasoning path toward answering the question.
  \item \textbf{Exploration Awareness}: Deprioritize candidates that resemble previously observed unproductive or low-yield exploration patterns.
  \item \textbf{Output Format}: The response must be a Python-parseable dictionary with candidate relation names as keys and scores as values.
\end{itemize}

\textbf{Example:}
\begin{itemize}[leftmargin=1.5em, itemsep=0pt, topsep=2pt]
  \item \textbf{Input:}
  \begin{itemize}
    \item \texttt{Question}: ``Where was the director of the movie Titanic born?''
    \item \texttt{Historical Path}: \{\texttt{movie.directed\_by}\}
    \item \texttt{Candidate Relations}: \{\texttt{person.place\_of\_birth}, \texttt{person.nationality}, \texttt{person.spouse}\}
    \item \texttt{Summary of Exploration Experience}: ``Exploration paths focusing on nationality are often too coarse when a specific birthplace is required, and spouse relations rarely contribute to location-based queries.''
  \end{itemize}
  \item \textbf{Output:}
\begin{verbatim}
{"person.place_of_birth": 0.9, "person.nationality": 0.2, "person.spouse": 0.1}
\end{verbatim}
\end{itemize}

\textbf{Your Task:}
\begin{itemize}[leftmargin=1.5em, itemsep=0pt, topsep=2pt]
  \item \texttt{Question}: \{question\}
  \item \texttt{Historical Path}: \{historical\_path\}
  \item \texttt{Candidate Relations}: \{top\_k\_relations\}
  \item \texttt{Summary of Exploration Experience}: \{exploration\_experience\}
\end{itemize}

\textbf{Output:}

\end{tcolorbox}

\captionof{figure}{Prompt template used in the Dual-Feedback Re-ranking module to re-rank candidate relations by integrating contextual coherence and prior exploration experience.}
\label{fig:prompt_reranking}

\end{document}